\documentclass[sigconf, nonacm]{acmart}
\AtBeginDocument{%
  }

\usepackage{graphicx}
\usepackage{subcaption}
\usepackage{hyperref}
\usepackage{wrapfig}
\usepackage{tcolorbox}
\usepackage{xcolor}
\usepackage{multirow}
\usepackage{colortbl}
\usepackage{makecell}

\setcopyright{acmlicensed}
\copyrightyear{2025}
\acmYear{2025}
\acmDOI{XXXXXXX.XXXXXXX}
\acmConference[Conference acronym 'XX]{}{}{}
\acmISBN{978-1-4503-XXXX-X/2018/06}

\begin{document}

\title{\textsc{SymLoc}: Symbolic Localization of Hallucination across HaluEval and TruthfulQA}




\author{Naveen Lamba}
\affiliation{%
  \department{CAIMIF \& Dept. of CSA}
  \institution{Sharda University}
  \city{Greater Noida}
  \country{India}
}
\email{naveenlamba30894@gmail.com}

\author{Sanju Tiwari}
\affiliation{%
  \department{CAIMIF \& Dept. of CSA}
  \institution{Sharda University}
  \city{Greater Noida}
  \country{India}
}
\email{tiwarisanju18@ieee.org}

\author{Manas Gaur}
\affiliation{%
  \institution{University of Maryland}
  \city{Baltimore County}
  \country{USA}
}
\email{manas@umbc.edu}

\renewcommand{\shortauthors}{}

\begin{abstract}
Large Language Models (LLMs) still struggle with hallucination, especially when confronted with symbolic triggers like modifiers, negation, numbers, exceptions, and named entities. Yet, we lack a clear understanding of where these symbolic hallucinations originate, making it crucial to systematically handle such triggers and localize the emergence of hallucination inside the model. While prior work explored localization using statistical techniques like Local Source Contribution (LSC) and activation variance analysis, these methods treat all tokens equally and overlook the role symbolic linguistic knowledge plays in triggering hallucinations. So far, no approach has investigated how symbolic elements specifically drive hallucination failures across model layers, nor has symbolic linguistic knowledge been used as the foundation for a localization framework.
We propose the first symbolic localization framework that leverages symbolic linguistic and semantic knowledge to meaningfully trace the development of hallucinations across all model layers. By focusing on how models process symbolic triggers—negation, named entities, numbers, and exceptions—we analyze five models using HaluEval and TruthfulQA. Our symbolic knowledge approach reveals that attention variance for these linguistic elements explodes to critical instability in early layers (2-4), with negation triggering catastrophic variance levels, demonstrating that symbolic semantic processing breaks down from the very beginning.
Through the lens of symbolic linguistic knowledge, despite larger model sizes, hallucination rates remain consistently high (78.3\%-83.7\% across Gemma variants), with steep attention drops for symbolic semantic triggers throughout deeper layers. Our findings demonstrate that hallucination is fundamentally a symbolic linguistic processing failure, not a general generation problem, revealing that symbolic semantic knowledge provides the key to understanding and localizing hallucination mechanisms in LLMs.
\end{abstract}




\maketitle

\section{Introduction}
Hallucination in Large Language Models (LLMs) fundamentally stems from failures in \textit{symbolic processing}---the model's ability to correctly interpret and manipulate abstract linguistic concepts that carry specific meaning, definitions, and symbolic constraints. In the context of LLMs, symbolic processing encompasses how models handle negation (which reverses truth values), named entities (which anchor factual claims), numerical values (which specify exact quantities), and logical exceptions (which create conditional reasoning paths). Symbolic features—such as negation, named entities, numerical values, modifiers, and exceptions—have been repeatedly associated with hallucination~\cite{Lamba2025}, since they impose greater burdens on a model's reasoning and factual grounding abilities, particularly when models need to manipulate abstract concepts or perform multi-step logical operations. If we can identify where semantic reasoning breaks down within model layers, we could develop targeted interventions to prevent hallucination at its source rather than detecting it post-hoc.

Consider a canonical example from TruthfulQA \cite{Lin2021}: ``\textit{Do we only use 10\% of our brains?}'' The correct answer requires logical interpretation to debunk this widespread myth: ``\textit{No, we use virtually all of our brain.}'' However, models frequently fail to engage the abstract reasoning required to negate false premises, instead generating hallucinated affirmations like ``\textit{Yes, Humans only utilize about 10\% of their brains at any given time}'' (as generated by Llama-3.1-8B). This failure demonstrates how models struggle with conceptual reasoning that requires contradicting common misconceptions. Similarly, HaluEval \cite{Li2023} questions involving specific named entities often trigger fabricated details when models encounter unfamiliar entity-attribute combinations.

\begin{figure*}[h]
    \centering
    \includegraphics[width=0.85\linewidth]{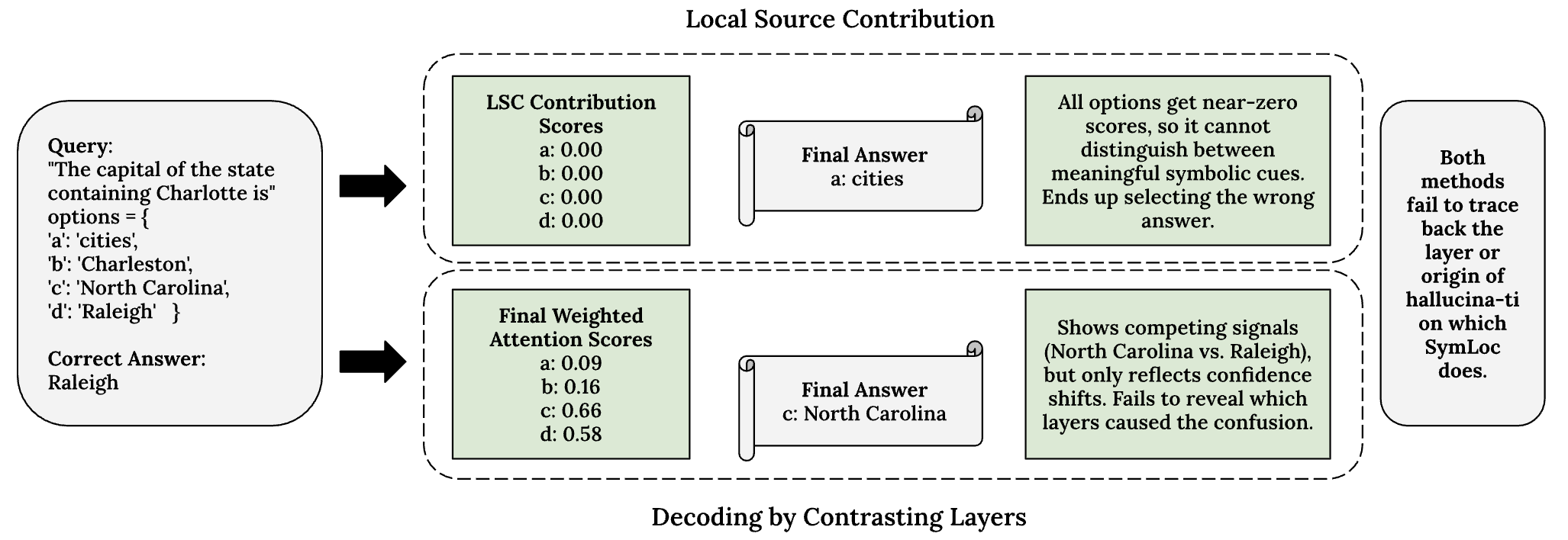}
    \caption{Comparison of LSC and DoLa on the query “The capital of the state containing Charlotte is.” LSC assigns flat scores and selects cities, while DoLa shows competing signals (North Carolina vs. Raleigh) but cannot pinpoint the layer of confusion. Both fail to localize the origin of hallucination; therefore, we did not consider them in our study.}
    \label{fig:placeholder}
\end{figure*}

\begin{figure}[h]
    \centering
    \includegraphics[width=1\linewidth]{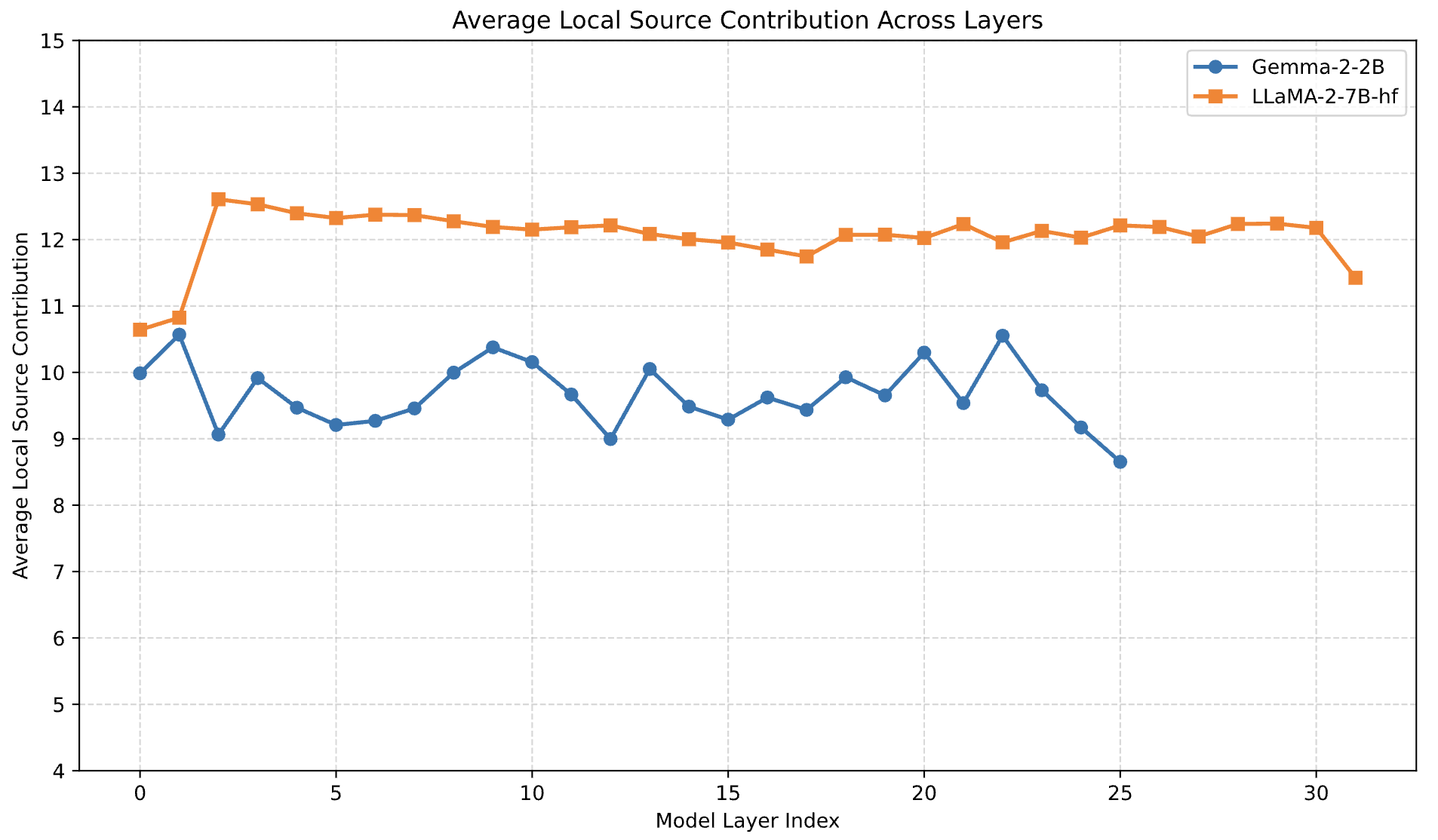}
    \caption{Average LSC across layers for Gemma-2-2B and Llama-2-7B-hf on HaluEval. Both models show flat or decaying attribution scores, failing to reflect symbolic instability or hallucination emergence.}
    \label{fig:lsc}
\end{figure}

Current hallucination localization methods face fundamental limitations in identifying when and where semantic interpretation fails. Local Source Contribution (LSC)~\cite{Xu2023} quantifies how much each input token contributes to generating specific output tokens through gradient-based attribution:
\begin{equation}
\text{LSC}(x_i, y_j) = \frac{\partial \log P(y_j|x_{1:i}, y_{1:j-1})}{\partial x_i}
\end{equation}
where $x_i$ represents the $i$-th input token and $y_j$ represents the $j$-th output token. However, as \autoref{fig:lsc} illustrates, LSC struggles to link attribution scores with the semantic origins of abstract concepts~\cite{Sriramanan2024}. Layer-wise approaches like Decoding by Contrasting Layers (DoLa)~\cite{Chuang2024} contrast logit shifts between later and earlier layers, enabling partial localization of factual knowledge. Yet, as shown in the example in \autoref{fig:placeholder}, DoLa surfaces competing candidates (e.g., Raleigh vs. North Carolina) without identifying the precise layer where reasoning collapses. Moreover, its outputs reveal conflicting signals across layers, blurring the distinction between genuine factual grounding and spurious noise. This suggests that while DoLa highlights tension between layers, it cannot deliver fine-grained explanations of how or when logical reasoning breaks down. Entropy-based methods~\cite{Kim2024} track information loss but cannot isolate breakdowns triggered by specific symbolic cues. Most critically, \textit{at which stage of the forward pass symbolic cues start to interfere with the model's internal representations remains unknown}.

To address these limitations, we propose \textbf{\textsc{SymLoc}}, a symbolic localization framework that traces the internal emergence of hallucination by focusing specifically on abstract concept manipulation failures. \textbf{\textsc{SymLoc}} introduces \textit{symbolic attention}---the attention models assign to tokens with symbolic properties---and tracks its variance across layers to detect when the processing of symbolic triggers becomes unstable. This approach moves beyond token-level attribution to diagnose hallucination as a structural phenomenon rooted in conceptual reasoning breakdowns.

Our key contributions are threefold. First, we develop the first framework to localize hallucination through the lens of semantic interpretation, identifying specific layers where abstract reasoning fails rather than treating all tokens equally. Second, we introduce symbolic attention variance as a novel metric that detects early encoding failures in logical processing, demonstrating that hallucinations originate in early layers (2--4) rather than late-stage decoding. Third, through systematic evaluation across five open-weight models on HaluEval and TruthfulQA benchmarks, we establish that symbolic attention variance peaks early and correlates with high hallucination rates, particularly for negation and named entities.

Our findings reveal that semantic reasoning failures emerge from early layer instability, with symbolic attention variance peaking in initial layers while hallucination rates remain persistently high across model scales. These results reframe hallucination as a layer-localized breakdown in abstract concept manipulation, providing the first interpretable approach to understanding when and where symbolic reasoning fails in LLMs.

\section{Related Work}


Hallucination detection in LLMs has been approached primarily as an output-level classification task. Methods like SelfCheckGPT \cite{Manakul2023} detect hallucinations in zero-resource, black-box settings by generating multiple outputs and checking for consistency across responses. Likewise, HaluEval \cite{Li2023} offers an extensive benchmark for detecting semantic hallucinations with annotations for truthfulness curated by humans. These techniques are post-hoc—they examine the produced text without investigating the underlying processes that lead to hallucination. This limits their ability to diagnose where or why hallucination arises in the model's computation graph. Recent work also shows that the internal state of LLMs can signal when they are likely to hallucinate \cite{Azaria2023}, but even these insights are not tied to localization within layers.

Efforts to reduce hallucination typically rely on prompt engineering, retrieval augmentation, or decoding interventions. For example, De-hallucination via formal methods proposes iterative prompting using formal logical constraints to steer outputs \cite{Jha2023}. Retrieval-Augmented Generation (RAG) approaches, such as Check Your Facts \cite{Peng2023} and Chain-of-Knowledge \cite{Li2024}, enhance factuality by grounding model responses in external knowledge. While these methods improve output correctness, they still treat hallucination as an emergent artifact of generation rather than as a structural failure distributed across the layers within the LLMs. Even mitigation lacks fine-grained control at the representational level.

An expanding collection of studies explores hallucination via the perspectives of symbolic reasoning and abstraction \cite{khandelwal2024domain}. For example, \citet{Zheng2023} emphasizes that hallucinations frequently arise when LLMs do not reason symbolically regarding logical elements such as negation or exceptions. Likewise, \citet{Su2024} recognizes entity mentions as a significant cause of symbolic misinterpretation, urging the modeling of semantically aware symbolic properties. However, these approaches do not systematically localize the problem across model layers; they focus on detecting symptoms rather than tracing internal breakdowns of symbolic encoding.

Mechanistic interpretability attempts to reverse-engineer neural models by identifying internal components, such as neurons, attention heads, or layers, responsible for specific behaviors. \citet{Lad2024} situates this approach as central to quantitative AI safety, highlighting that identifying causal circuits in models enables more targeted intervention and reliable analysis of failure modes, especially in tasks that demand factuality and truthfulness. Building further on this, \citet{Garca-Carrasco2024} outlines a framework where mechanistic analysis is used to identify and learn about vulnerabilities in LLMs and demonstrate that hallucinations can stem from certain subcomponents or layers that misrepresent symbolic or factual inputs. Such representational failures, albeit enlightening, are model-specific and fail to generalize across model types and sizes.

While both activations and attention weights have been used in mechanistic interpretability, our study prioritizes attention for symbolic localization. Activation-based methods, such as patching or variance analysis, provide useful signals but are high-dimensional and difficult to link directly to symbolic cues \cite{jin2024mechanistic}. In contrast, attention explicitly encodes how models allocate focus to input tokens, offering a more interpretable proxy for whether symbolic elements—such as negation or exceptions—are consistently represented. Prior work also shows that unstable attention over semantically critical tokens often precedes factual errors, reinforcing attention’s diagnostic value for hallucination analysis \cite{Zheng2024, joshi2024towards, mohammadi2024welldunn}.

Building on this, our contribution introduces a symbolic localization framework that measures attention variance over symbolic cues across layers. This enables scalable, layer-wise diagnosis of hallucination grounded in mechanistic reasoning. By focusing on symbolic triggers, we move beyond surface-level output inspection and instead identify the structural causes of hallucination within the model’s internal computation.

\begin{table*}[!h]
\centering
\caption{Symbolic hallucination percentage statistics for original QA task across model sizes and datasets.}
\scriptsize
\begin{tabular}{l|cc|cc|cc|cc|cc}
\toprule
\multirow{2}{*}{\textbf{Symbolic Property}} & \multicolumn{2}{c|}{\textbf{Gemma-2-2B}} & \multicolumn{2}{c|}{\textbf{Gemma-2-9B}} & \multicolumn{2}{c}{\textbf{Gemma-2-27B}} & \multicolumn{2}{c}{\textbf{Llama-2-7B-hf}} & \multicolumn{2}{c}{\textbf{Llama-3.1-8B}} \\
 & HaluEval & TruthfulQA & HaluEval & TruthfulQA & HaluEval & TruthfulQA & HaluEval & TruthfulQA & HaluEval & TruthfulQA \\
\midrule
Modifiers         & 84.42 & 99.74  & 79.44 & 99.22 & 77.24 & 86.19 & 73.31 & 90.93 & 82.55 & 91.45 \\
Named Entities    & 83.58 & 99.69 & 78.52 & 99.38 & 76.43 & 88.46 & 73.09 & 93.85 & 81.85 & 94.46 \\
Numbers           & 85.38 & 100 & 80.42 & 100 & 76.32 & 94.00 & 76.76 & 91.04 & 81.98 & 92.54 \\
Negation          & 82.29 & 99.07 & 74.86 & 99.07 & 80.00 & 95.83 & 70.29 & 95.33 & 81.14 & 96.26 \\
Exceptions        & 90.91 & 100 & 81.82 & 100 & 80.00 & 90.74 & 76.77 & 94.57 & 83.84 & 93.48 \\
\bottomrule
\end{tabular}
\label{tab:combined_QA_symbolic}
\end{table*}

\section{\textsc{SymLoc} Methodology}
\begin{figure*}[!h]
    \centering
    \includegraphics[width=0.8\linewidth]{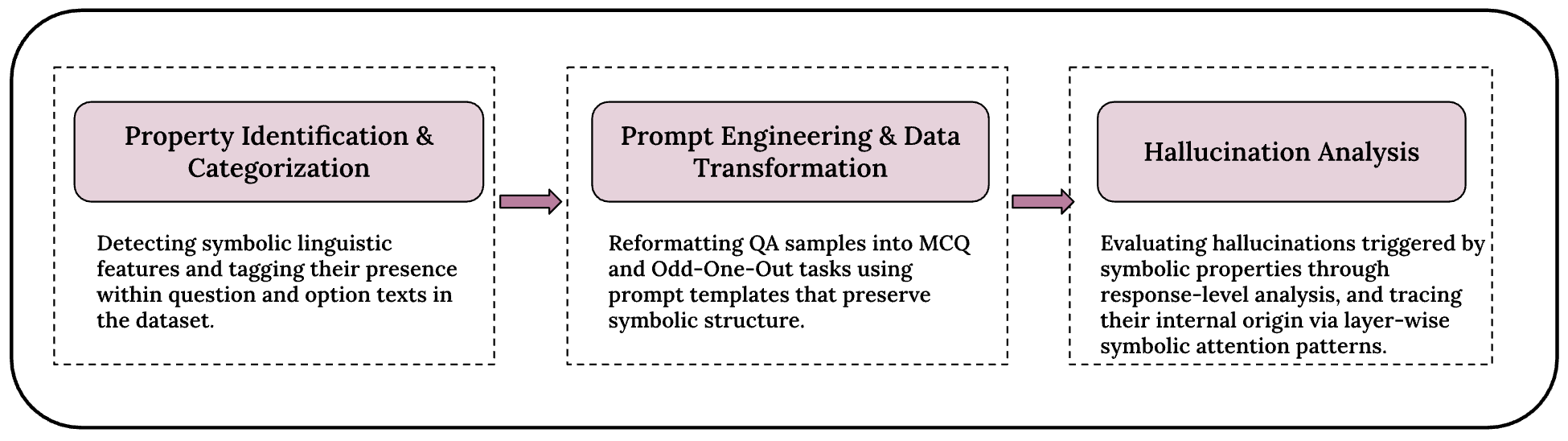}
    \caption{Structured evaluation pipeline. Benchmarks (HaluEval and TruthfulQA) are symbolically annotated and reformulated into QA, MCQ, and odd-one-out tasks. Models are then evaluated with attention analysis across layers to trace the origin of hallucinations}.    
    \label{fig:methodolody}
\end{figure*}

Our methodology consists of three main components, summarized in \autoref{fig:methodolody}: (i) dataset preparation and task conversion, (ii) symbolic property identification, and (iii) hallucination evaluation strategy. This structured pipeline allows us to systematically analyze how symbolic triggers destabilize model reasoning and trace where symbolic instability emerges within the model.

\subsection{Dataset Preparation and Task Conversion}

We evaluate five open-weight models—Gemma-2B, 9B, and 27B (Google DeepMind) \cite{Team2024}, and Llama-2–7B-hf and Llama-3.1-8B (Meta) \cite{Team2024l}—using two widely adopted hallucination benchmarks: HaluEval (1000 samples) and TruthfulQA (817 samples). Both datasets contain factual question-answer pairs, which we directly use as ground truth for evaluating hallucination. To robustly test hallucination under varying reasoning demands, we convert each dataset into three formats: (i) QA Format: Open-ended structure, exposing models to maximal freedom and highest hallucination risk, (ii) MCQ Format: Items reformulated into multiple-choice questions with one correct and two distractors, restricting generative freedom, and (iii) Odd-One-Out (OOO) Format: Lists conceptually related items with one semantic outlier, testing reasoning beyond factual recall. This multi-format evaluation allows us to assess hallucination consistently across open-ended, constrained, and comparative reasoning tasks.  

This results in a symbolic hallucination testbed of 5451 input instances (1817 samples × 3 formats). Each transformation preserves symbolic cues while altering the format to isolate hallucination behavior under different cognitive demands. To ensure consistency and eliminate sampling variability, we use each model’s recommended decoding settings (typically low or zero temperature) \cite{Liu2023evalplus}. Standardized prompts were used for each format:

\begin{itemize}
    \item \textbf{QA Prompt: }
    \begin{tcolorbox}[boxrule=0.3pt, colback=gray!5!white, colframe=gray!60!black, left=2pt, right=2pt, top=1pt, bottom=1pt]
    "Answer the following question in one short, factual sentence"
    \end{tcolorbox}

    \item \textbf{MCQ Prompt: }
    \begin{tcolorbox}[boxrule=0.3pt, colback=gray!5!white, colframe=gray!60!black, left=2pt, right=2pt, top=1pt, bottom=1pt]
    "You are a multiple-choice quiz solver. Read the question and select only the correct option (A, B, or C)"
    \end{tcolorbox}

    \item \textbf{Odd One Out Prompt: }
    \begin{tcolorbox}[boxrule=0.3pt, colback=gray!5!white, colframe=gray!60!black, left=2pt, right=2pt, top=1pt, bottom=1pt]
    "You are an expert in reasoning and comparison. Your task is to identify the odd one out from a list of three options. Only one option is unrelated to the others. Clearly state the odd option and explain why it is different."
    \end{tcolorbox}
\end{itemize}

These templates reduce ambiguity, ensuring hallucinations are attributable to internal reasoning failures rather than prompt artifacts.

\subsection{Property Identification and Categorization}
Each input was annotated with one or more of five symbolic properties known to challenge language models. Properties were identified using rule-based extraction, POS tagging, NER, dependency parsing, and manual verification for contextual accuracy. All automatic analyses were performed using \texttt{spaCy} with the \textit{en\_core\_web\_sm} model.  Below, we outline each property with examples, its role in hallucination and the method used to detect it in the samples:

\begin{enumerate}
    \item \textbf{Modifiers (adjectives, adverbs, descriptive verbs):} These components provide subjective or comparative details that enhance interpretative latitude. \textit{Example: “Which is the least controversial reform introduced in the past decade?”}. Words like “least” or “controversial” prompt nuanced judgments. Words like least or controversial trigger unverifiable assertions, increasing hallucination risk. To identify these modifiers, each sentence was analyzed using POS tagging, and tokens with tags \texttt{ADJ} (Adjectives), \texttt{ADV} (Adverbs), or \texttt{VERB} (Verbs) were flagged.
    \item \textbf{Named Entities (people, places, organizations):} This includes proper nouns that require grounding in external factual knowledge. \textit{Example: “What position does Dr. Elena Foster hold at NovaGen Institute?”}. Rare or unfamiliar entities often cause hallucinated roles, affiliations, or identities. These entities were extracted using Named Entity Recognition (NER), identifying tokens labeled as \texttt{PERSON} (Person), \texttt{ORG} (Organization), \texttt{GPE} (Geopolitical entity), or \texttt{LOC} (Location).
    \item \textbf{Numbers (quantitative expressions, dates, counts):} This category includes cardinal numbers, measurements, dates, and quantities. \textit{Example: “In what year did the Andean Treaty come into force?”}. Models often fabricate numeric details when data is inconsistent or missing. Numbers were identified by checking tokens for the \texttt{like\_num} attribute, allowing systematic detection of all numeric references in the samples. 
    \item \textbf{Negation (logical operators such as “not”, “never”, “none”):} This negation inverts the expected logical structure of a sentence. \textit{Example: “Which of these countries is not a member of the Arctic Council?”}. Negation cues are often ignored, leading to logically inconsistent outputs. Identification was performed by matching tokens against a predefined set of negation keywords, including “not,” “never,” “none,” “cannot,” and several others, ensuring that all negation cues were captured in context.
    \item \textbf{Exceptions (rule violaters, rare cases, uncommon scenarios):} These inputs evaluate the model’s capacity to surpass generalizations or manage counterexamples. \textit{Example: “Which mammal lays eggs instead of giving birth to live offspring?”}. Exceptions demand contextual understanding beyond statistical regularities. Models often default to the common case rather than recognizing factual outliers. Exception cues were identified by searching for keywords such as “except,” “but,” “excluding,” “however,” and “although,” which mark deviations from the general pattern, supplemented by manual review to confirm rare-case contexts.
\end{enumerate}

By categorizing prompts according to these symbolic properties, we aim to isolate symbolic triggers that systematically correlate with hallucination across different tasks, prompt formats, and model scales. This symbolic framework enables a controlled investigation into the semantic and syntactic weaknesses underlying hallucination behavior in LLMs.

\subsection{Hallucination Localization Strategy}

We apply a two-tier evaluation strategy: symbolic trigger-driven hallucination scoring and analysis of layer-wise representational variance.

\subsubsection{Symbolic Trigger-Based Hallucination Percentage}

We compute the hallucination frequency for each symbolic property by evaluating model outputs corresponding to inputs that contain a specific symbolic trigger. A symbolic hallucination occurs when the model produces a confidently incorrect output in response to an input annotated with that symbolic property. The hallucination percentage for a symbolic category $S$ is defined as:

{\footnotesize
\begin{equation}
\text{Hallucination}_{\text{S}} = 
\frac{\text{\ hallucinated outputs for inputs with } S}
     {\text{\ total inputs containing } S} \times 100
\end{equation}
}

This metric quantifies how often the presence of a symbolic trigger in the input leads to hallucinated responses, and is computed independently for each symbolic property across all models and task formats.

\subsubsection{Layer-Wise Symbolic Attention Analysis}

To understand how symbolic properties are handled within model layers, we compute the median attention received by symbolic tokens at each transformer layer. For each symbolic category $S$, we first identify its corresponding tokens in the input sequence using part-of-speech tags, named entity recognition, and keyword lists.

Let $T$ denote the number of tokens in the input sequence. For each layer $l$ and head $h$, the attention matrix is represented as $A_l^{(h)} \in \mathbb{R}^{T \times T}$. Let $\mathcal{T}_S$ be the set of token positions corresponding to symbolic property $S$. We calculate the average attention received by symbolic tokens at layer $l$ as:

{\footnotesize
\begin{equation}
a_l^{(S)} = \frac{1}{|H| \cdot |T| \cdot |\mathcal{T}_S|} \sum_{h=1}^{H} \sum_{i=1}^{T} \sum_{j \in \mathcal{T}_S} A_l^{(h)}[i, j]
\end{equation}
}

This process is repeated across four independent iterations\footnote{Independent iterations implies that each iteration is executed in isolation, with the model reset prior to execution. Thus, no information regarding generations or answers from earlier iterations is accessible in subsequent runs.} to ensure robustness. For each symbolic category and each layer, we report the \textit{median} of all collected attention values:

{\footnotesize
\begin{equation}
\text{MedianAttention}_l^{(S)} = \text{median}(\{ a_l^{(S)} \}_{\text{samples, iter}})
\end{equation}
}

We also compute the \textit{standard deviation} across the same values to quantify variability:

{\footnotesize
\begin{equation}
\label{eq:std}
\text{SD}_l^{(S)} = \sqrt{\frac{1}{N} \sum_{k=1}^{N} \left(a_{l,k}^{(S)} - \mu_l^{(S)} \right)^2}
\end{equation}
}

where $N$ is the total number of symbolic attention values collected for layer $l$ and property $S$, and $\mu_l^{(S)}$ is the sample mean.

This layer-wise symbolic attention profiling helps identify where in the model symbolic information is concentrated, and how consistently it is attended to across samples. High median attention in early layers may suggest symbolic salience, while high variance could signal representational instability associated with hallucination.

These metrics enable us to trace how symbolic properties are weighted across model depth, and identify early-layer patterns associated with hallucination onset.

In deep learning, inconsistent attention over semantically meaningful tokens—such as negation cues or exception clauses—can lead to fragmented internal representations that degrade the quality of encoding and impair factual grounding. When symbolic cues that encode logical structure (e.g., “not”, “unless”, named entities) are attended to unevenly across layers, the model may struggle to build coherent meaning, increasing the likelihood of hallucinated outputs. Research in mechanistic interpretability has shown that unstable attention patterns over such critical tokens often precede factual generation errors, indicating that symbolic attention variance may not simply correlate with hallucination, but reflect a deeper representational failure within the model’s internal circuitry \cite{Garca-Carrasco2024, Lad2024}.

\section{Results and Analysis}

We organize our results around four central themes: Symbolic Hallucination Persists Across Model Scales, Symbolic Triggers Robust to Input Length, Attention-Level Traces of Symbolic Instability, and Symbolic Localization Across Model Layers.

\subsection{Symbolic Hallucination Persists Across Model Scales}

In the QA format, modifiers, named entities, and negation consistently emerge as the most hallucination-prone symbolic properties across both Gemma and Llama models. As shown in \autoref{tab:combined_QA_symbolic}, Gemma-2-2B exhibits high hallucination rates for modifiers (84.76\%), named entities (83.87\%), and negation (82.29\%) on the HaluEval dataset. These rates decline only slightly with scale—dropping to 77.24\%, 76.43\%, and 80.00\%, respectively, in Gemma-2-27B—indicating that scaling alone offers limited improvement.

This trend holds for TruthfulQA as well, where symbolic hallucination remains elevated across all model sizes. Modifiers peak at 94.98\% in Gemma-2-9B, while negation remains above 95\% across the board (e.g., 99.07\% in both Gemma-2-2B and 9B). Even categories with fewer instances, such as exceptions, show stubbornly high hallucination rates in TruthfulQA (e.g., 95.45\% in Gemma-2-2B).

A similar pattern is observed in the Llama series. Llama-3.1–8B records symbolic hallucination rates above 80\% for modifiers, named entities, and negation in both datasets, with negligible reduction compared to smaller models.

\subsection{Symbolic Triggers Robust to Input Length}

\begin{table*}[!h]
\centering
\caption{Hallucination \% by symbolic property and question token length across Gemma and Llama models and datasets (HaluEval, TruthfulQA). For HaluEval, hallucination values for the \textbf{exception} property are 0 in all length bins except 10–19 and 30–39 due to limited occurrences. For TruthfulQA, hallucination \% is 0 in the 50+ length bin, as all questions were shorter than 50 tokens.} 

\scriptsize
\begin{tabular}{c|c|cc|cc|cc|cc|cc}
\toprule
\rowcolor{gray!30}\textbf{Query Token} & \textbf{Symbolic} & \multicolumn{2}{c|}{\textbf{Gemma-2-2B}} & \multicolumn{2}{c|}{\textbf{Gemma-2-9B}} & \multicolumn{2}{c}{\textbf{Gemma-2-27B}} & \multicolumn{2}{c}{\textbf{Llama-2-7B-hf}} & \multicolumn{2}{c}{\textbf{Llama-3.1-8B}} \\
 \rowcolor{gray!30}\textbf{Length}& \textbf{Property} & HaluEval & TruthfulQA & HaluEval & TruthfulQA & HaluEval & TruthfulQA & HaluEval & TruthfulQA & HaluEval & TruthfulQA \\
\midrule
\multirow{5}{*}{0–9}
 & \cellcolor{gray!30}Modifiers        & 63.73 & 90.07 & 65.69 &	90.07 &	60.78 &	90.07 & 56.86 &	80.60 &	68.63 &	81.06 \\
 & \cellcolor{gray!30}Named Entities   & 49.02 &	29.33 &	47.06 &	29.33 &	46.08 &	29.33 & 45.10 &	27.02 &	54.90 &	27.25 \\
 & \cellcolor{gray!30}Numbers          & 10.78 &	5.77 &	9.80 &	5.77 &	9.80 &	5.77 & 9.80 &	4.85 &	10.78 &	5.08 \\
 & \cellcolor{gray!30}Negation         & 1.96 &	7.16 & 	0.98 &	7.16 &	1.96 &	7.16 & 0.98 &	6.47 &	2.94 &	6.47 \\
 & \cellcolor{gray!30}Exceptions       & 0.98 &	9.01 &	0.98 &	9.01 &	0.98 &	9.01 & 0.98 &	8.55 &	0.98 &	8.08\\
\midrule
\multirow{5}{*}{10–19}
 & \cellcolor{gray!30}Modifiers        & 83.78 &	98.77 &	79.10 &98.77 & 78.26 & 99.38& 73.75 & 92.92 &	82.61 &	92.92 \\
 & \cellcolor{gray!30}Named Entities   & 66.22 &	45.54 &	63.04 &	45.85 &	61.20 &	45.85 & 58.53 & 44.31 &	65.22 &	43.69 \\
 & \cellcolor{gray!30}Numbers          & 26.59 &	7.08 &	25.75 &	7.08 &	24.41 &	7.08 & 24.58 & 6.77 & 26.42 & 6.77 \\
 & \cellcolor{gray!30}Negation         & 11.54 &	16.31 &	11.04 &	16.62 &	11.04 &	16.62 & 10.37 &	16.62 &	11.71 &	16.62 \\
 & \cellcolor{gray!30}Exceptions       & 7.53 &	13.54 &	7.36 &	13.54 &	7.36 &	13.54 & 6.86 &	12.62 &	7.36 &	12.92\\
\midrule
\multirow{5}{*}{20–29}
 & \cellcolor{gray!30}Modifiers        & 83.25 &	100.00 & 76.85 & 95.56 & 79.31 & 97.78 & 72.91 & 93.33 & 80.79 & 95.56\\
 & \cellcolor{gray!30}Named Entities   & 76.85 &	88.89 &	70.94 &	84.44 &	73.40 &	86.67 & 68.47 & 82.22 & 74.88 & 86.67 \\
 & \cellcolor{gray!30}Numbers          & 48.77 &	33.33 &	45.32 &	33.33 &	46.31 &	33.33 & 44.33 &	33.33 &	46.31 &	33.33\\
 & \cellcolor{gray!30}Negation         & 23.15 &	40.00 &	20.69 &	37.78 &	19.70 &	37.78 & 19.70 &	35.56 &	22.17 &	37.78\\
 & \cellcolor{gray!30}Exceptions       & 13.30 &	15.56 &	11.33 &	15.56 &	10.84 &	15.56 & 10.84 & 15.56 &	12.81 &	15.56\\
\midrule
\multirow{5}{*}{30–39}
 & \cellcolor{gray!30}Modifiers        & 81.13 &	100.00 & 73.58 & 90.00 & 69.81 & 90.00 & 64.15 & 70.00 & 71.70 & 80.00\\
 & \cellcolor{gray!30}Named Entities   & 77.36 &	70.00 &	69.81 &	70.00 &	66.04 &	70.00 & 60.38 & 60.00 &	67.92 &	70.00\\
 & \cellcolor{gray!30}Numbers          & 62.26 &	20.00 &	58.49 &	20.00 &	56.60 &	20.00 & 50.94 &	20.00 &	56.60 &	20.00\\
 & \cellcolor{gray!30}Negation         & 28.30 &	40.00 &	22.64 &	40.00 &	24.53 & 40.00 & 20.75 &	40.00 &	24.53 &	40.00\\
 & \cellcolor{gray!30}Exceptions       & 15.09 &	20.00 &	13.21 &	20.00 &	11.32 &	20.00 & 13.21 &	20.00 &	13.21 &	20.00\\
\midrule
\multirow{5}{*}{40–49}
 & \cellcolor{gray!30}Modifiers        & 77.78 &	100.00 & 59.26 & 66.67 & 62.96 & 66.67 & 48.15 & 33.33 & 59.26 & 33.33\\
 & \cellcolor{gray!30}Named Entities   & 74.07 &	33.33 &	59.26 &	33.33 &59.26 &	33.33 & 48.15 &	0.00 &	59.26 &	0.00\\
 & \cellcolor{gray!30}Numbers          & 51.85 &	33.33 &	40.74 &	33.33 &	44.44 &	33.33 & 37.04 &	0.00 &	44.44 &	0.00\\
 & \cellcolor{gray!30}Negation         & 18.52 &	0.00 &	14.81 &	0.00 &	18.52 &	0.00 & 14.81 &	0.00 &	18.52 &	0.00\\
 & \cellcolor{gray!30}Exceptions       & 22.22 &	0.00 &	14.81 &	0.00 &	11.11 &	0.00 & 11.11 &	0.00 &	14.81 &	0.00\\
\midrule
\multirow{5}{*}{50+}
 & \cellcolor{gray!30}Modifiers        & 82.35 &	100.00 & 82.35 & 100.00 & 88.24 & 100.00 & 70.59 & 100.00 & 76.47 & 100.00\\
 & \cellcolor{gray!30}Named Entities   & 82.35 &	100.00 & 82.35 & 100.00	 & 88.24 & 100.00 & 70.59 & 100.00 &	76.47 &	100.00\\
 & \cellcolor{gray!30}Numbers          & 64.71 &	100.00 & 58.82 & 100.00 & 64.71 & 100.00 & 58.82 & 100.00 & 52.94 & 100.00\\
 & \cellcolor{gray!30}Negation         & 35.29 &	0.00 & 35.29 & 0.00 & 41.18 & 0.00 & 9.41 & 0.00 & 35.29 & 0.00\\
 & \cellcolor{gray!30}Exceptions       & 17.65 &	0.00 & 11.76 & 0.00 & 17.65 & 0.00 & 11.76 & 0.00 & 5.88 & 0.00\\
\bottomrule
\end{tabular}
\label{tab:symbolic_to_length}
\end{table*}

To understand whether symbolic hallucination varies with query complexity, we analyze how hallucination rates change across different input lengths. \autoref{tab:symbolic_to_length} shows average hallucination percentages for each symbolic property—modifiers, named entities, numbers, negation, and exceptions—grouped by token length brackets. 

We observe that hallucination rates remain consistently high across input lengths, with modifiers and named entities being especially vulnerable. For instance, modifier-triggered hallucinations peak at 96.87\% for short inputs (0–29 tokens) and remain above 88\% in the 30–39 token range. Named entities follow a similar trend, reaching nearly 78\% in short queries—lengths that closely resemble everyday user inputs. Although longer inputs (40+ tokens) occasionally exhibit reduced hallucination, likely due to the availability of richer context, the effect is inconsistent, particularly for negation and exceptions. Reported values of 0\% indicate the absence of a given symbolic property in that bin rather than true model accuracy. Taken together, these results show that symbolic hallucination persists irrespective of input length, suggesting that its origin lies in how symbolic cues are internally encoded within the model rather than being a simple artifact of brevity.

\subsection{Attention-Level Traces of Symbolic Instability}

To understand how task format impacts symbolic representation, we analyze symbolic token attention across QA, MCQ, and Odd-One-Out (OOO) prompts using the Gemma and Llama model families. \autoref{tab:attention_symbolic} reports average attention to symbolic tokens at mid-to-late layers. For layer selection, we follow prior work highlighting the importance of mid-to-late transformer layers for semantic integration and context sensitivity \cite{Wu2025, Liu2023} \footnote{Layer indices are adopted from prior studies focusing on internal transformer layers. In our analysis, these layers provide insight into how symbolic cues are heuristically encoded. However, we additionally examined earlier layers and found that symbolic hallucinations first tend to emerge there.}. Accordingly, we probed Layers 10 and 20 for Gemma-2-2B, Layers 20 and 31 for Gemma-2-9B,  Layers 23 and 40 for Gemma-2-27B, Layers 15 and 28 for Llama models. This consistency across scales enables a meaningful comparison of symbolic attention dynamics.

Across models and symbolic properties, QA consistently receives higher symbolic attention than MCQ and OOO. For example, in Gemma-2-27B, attention to modifiers drops from 0.0078 in QA to 0.0063 in MCQ, and slightly rises to 0.0085 in OOO. Similarly, named entity attention is highest in QA (0.0165), lower in MCQ (0.0116), and lowest in OOO (0.0106). This pattern holds across other properties like negation and exceptions. Llama models follow similar trends. In Llama-3.1–8B, attention to negation drops from 0.0183 in QA to 0.0155 in MCQ and 0.0083 in OOO. MCQ prompts also show the steepest attention decay across layers, suggesting reduced symbolic retention under more constrained formats. These results suggest that symbolic instability is task-sensitive—with QA encouraging stronger symbolic grounding, while MCQ and OOO dilute symbolic focus, potentially amplifying hallucination risk in deeper layers. 

To further probe where symbolic cues begin to weaken, we analyze \autoref{tab:layer_localization}, which contrasts the attention assigned to symbolic tokens against the highest-attended token at the same layer. Strikingly, across Gemma and Llama variants, the first instability appears consistently in the early layers (L3–L4), irrespective of property type. At these layers, we observe a sudden drop in attention to symbolic tokens, with their weights overshadowed by more generic or functional tokens. For instance, in the negation example (“Which of these countries is not a member of the Arctic Council?”), the negation cue “not” receives less attention (0.0451) than the interrogative “which” (0.1152). Similarly, for named entities, the symbolic sub-token with the highest attention (e.g., “Elena”) is still deprioritized relative to surrounding context tokens such as “position” (0.0938 in Gemma-2-2B) or even punctuation (0.3041 in Llama-2-7B-hf). Comparable drops occur for modifiers (“least” vs. “which”), numbers (“year” vs. “in”), and exceptions (“instead” vs. “giving”), showing that logically and factually critical cues lose dominance almost immediately after embedding.

This early overshadowing suggests that symbolic instability originates during shallow encoding rather than later decoding, reinforcing our broader finding that hallucinations are structurally rooted in representational breakdowns of symbolic cues. Consistent with prior work on mechanistic interpretability, where unstable attention in early layers has been linked to factual errors \cite{jin2024mechanistic, rogers2020primer}, our results position early symbolic variance as a key diagnostic signal for hallucination onset. We build on this observation in Section 4.4 by tracing layer-wise symbolic attention variance, showing how early instability propagates forward through the model.

\begin{table*}[!h]
\centering
\caption{Attention of symbolic tokens at different layers for QA task across model sizes and datasets.}
\scriptsize
\begin{tabular}{c|l|cc|cc|cc|cc|cc}
\toprule
\multirow{2}{*}{\textbf{Task}} &\multirow{2}{*}{\textbf{Symbolic Property}} & \multicolumn{2}{c|}{\textbf{Gemma-2-2B}} & \multicolumn{2}{c|}{\textbf{Gemma-2-9B}} & \multicolumn{2}{c}{\textbf{Gemma-2-27B}} & \multicolumn{2}{c}{\textbf{Llama-2-7B-hf}} & \multicolumn{2}{c}{\textbf{Llama-3.1-8B}} \\
 & & Layer 10 & Layer 20 & Layer 20 & Layer 31 & Layer 23 & Layer 40 & Layer 15  & Layer 28  & Layer 15  & Layer 28  \\
\midrule
\multirow{5}{*}{QA} & Modifiers         & 0.0100 & 0.0097 & 0.0095 & 0.0092 & 0.0078 & 0.0059 & 0.0058 & 0.0015 & 0.0094 & 0.0034 \\
                    & Named Entities    & 0.0147 & 0.0082 & 0.0168 & 0.0095 & 0.0165 & 0.0063 & 0.0132 & 0.0033 & 0.0175 & 0.0032 \\
                    & Numbers           & 0.0114 & 0.0056 & 0.0122 & 0.0060 & 0.0117 & 0.0047 & 0.0089 & 0.0082 & 0.0137 & 0.0018 \\
                    & Negation          & 0.0172 & 0.0091 & 0.0182 & 0.0070 & 0.0137 & 0.0062 & 0.0084 & 0.0030 & 0.0183 & 0.0031 \\
                    & Exceptions        & 0.0166 & 0.0118 & 0.0134 & 0.0101 & 0.0158 & 0.0072 & 0.0076 & 0.0048 & 0.0120 & 0.0031 \\
\midrule

\multirow{5}{*}{MCQ}& Modifiers         & 0.0093 & 0.0068 & 0.0084 & 0.0067 & 0.0063 & 0.0039 & 0.0046 & 0.0011 & 0.0078 & 0.0024 \\
                    & Named Entities    & 0.0177 & 0.0051 & 0.0134 & 0.0062 & 0.0116 & 0.0040 & 0.0098 & 0.0021 & 0.0123 & 0.0021 \\
                    & Numbers           & 0.0104 & 0.0038 & 0.0095 & 0.0043 & 0.0085 & 0.0030 & 0.0203 & 0.0216 & 0.0106 & 0.0012\\
                    & Negation          & 0.0206 & 0.0067 & 0.0147 & 0.0052 & 0.0103 & 0.0042 & 0.0077 & 0.0021 & 0.0155 & 0.0022 \\
                    & Exceptions        & 0.0140 & 0.0083 & 0.0107 & 0.0072 & 0.0121 & 0.0050 & 0.0056 & 0.0032 & 0.0093 & 0.0022\\
\midrule

\multirow{5}{*}{OOO}& Modifiers         & 0.0076 & 0.0071  & 0.0077 & 0.0062 & 0.0085 & 0.0040 & 0.0049 & 0.0017 & 0.0087 & 0.0022 \\
                    & Named Entities    & 0.0087 & 0.0055 & 0.0123 & 0.0055 & 0.0106 & 0.0035 & 0.0052 & 0.0015 & 0.0093 & 0.0019 \\
                    & Numbers           & 0.0050 & 0.0031 & 0.0074 & 0.0032 & 0.0063 & 0.0052 & 0.0187 & 0.0196 & 0.0053 & 0.0017 \\
                    & Negation          & 0.0092 & 0.0068 & 0.0084 & 0.0054 & 0.0082 & 0.0035 & 0.0074 & 0.0014 & 0.0083 & 0.0016 \\
                    & Exceptions        & 0.0072 & 0.0047 & 0.0070 & 0.0064 & 0.0085 & 0.0035 & 0.0055 & 0.0016 & 0.0102 & 0.0016 \\
\bottomrule
\end{tabular}
\label{tab:attention_symbolic}
\end{table*}

\subsection{Symbolic Localization Across Model Layers}

We analyze the layer-wise origin of hallucinations in transformer-based LLMs by tracking the standard deviation of attention over symbolic tokens. The metric defined in Equation 5 quantifies how inconsistently models attend to categories such as negation, modifiers, exceptions, numbers, and named entities across layers. Elevated variance is interpreted as representational instability, a potential precursor to hallucination.

\begin{figure*}[t]
  \centering
  \begin{subfigure}[b]{0.48\textwidth}
    \includegraphics[width=\textwidth]{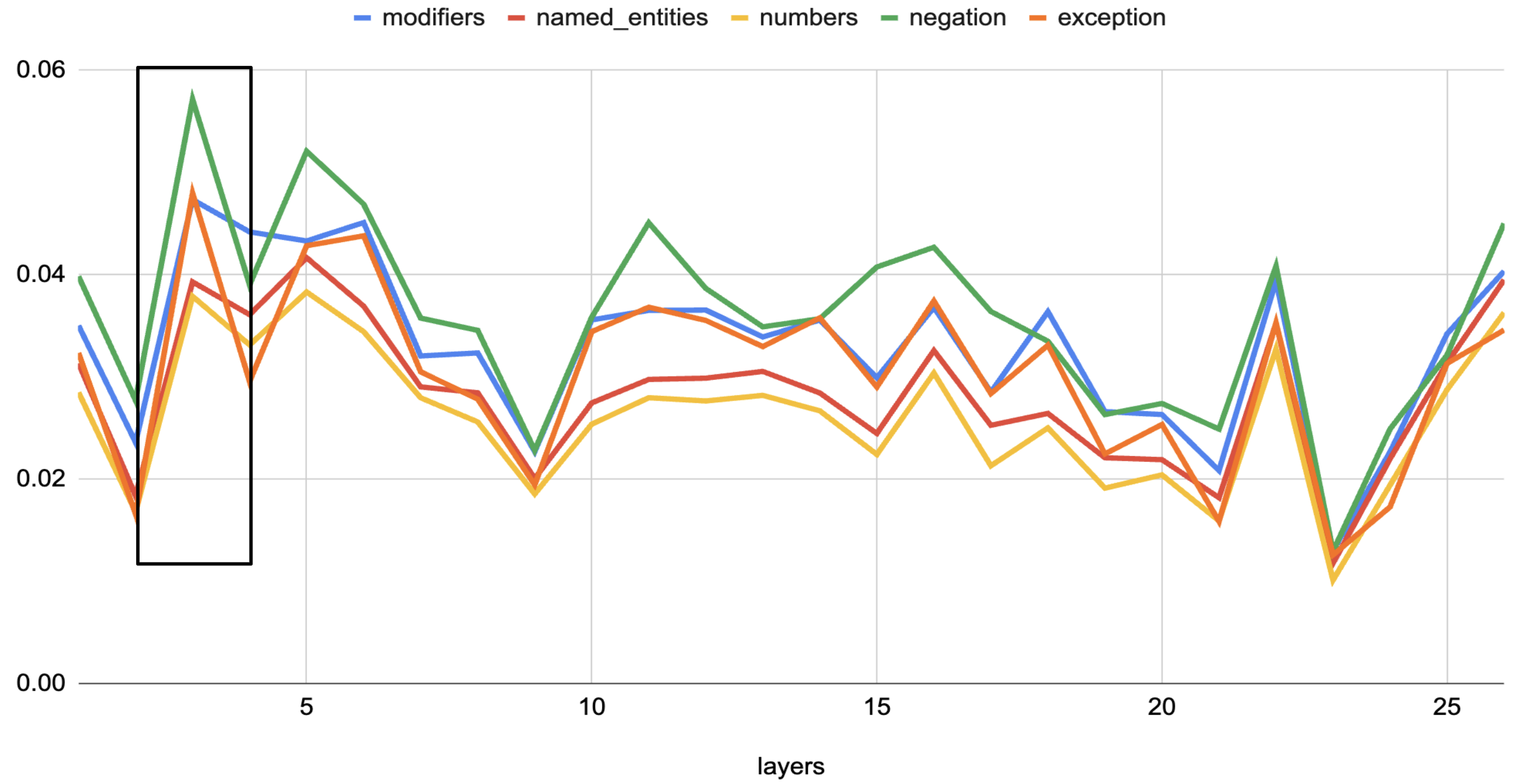}
    \caption{Gemma-2-2B (Total Layers-26)}
    \label{fig:gemma2b}
  \end{subfigure}
  \hfill
  \begin{subfigure}[b]{0.48\textwidth}
    \includegraphics[width=\textwidth]{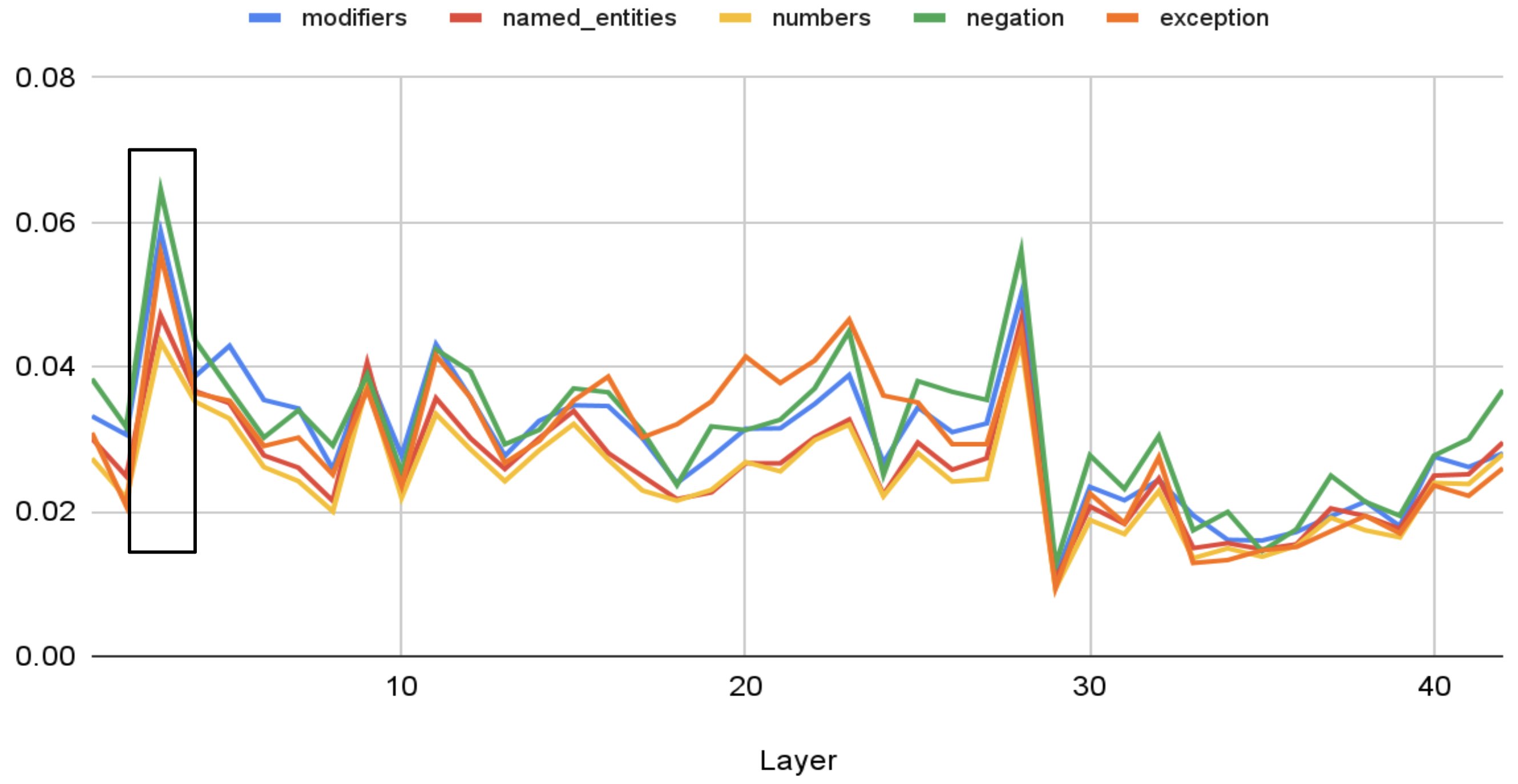}
    \caption{Gemma-2-9B (Total Layers-42)}
    \label{fig:gemma9b}
  \end{subfigure}
  \hfill
  \begin{subfigure}[b]{0.48\textwidth}
    \includegraphics[width=\textwidth]{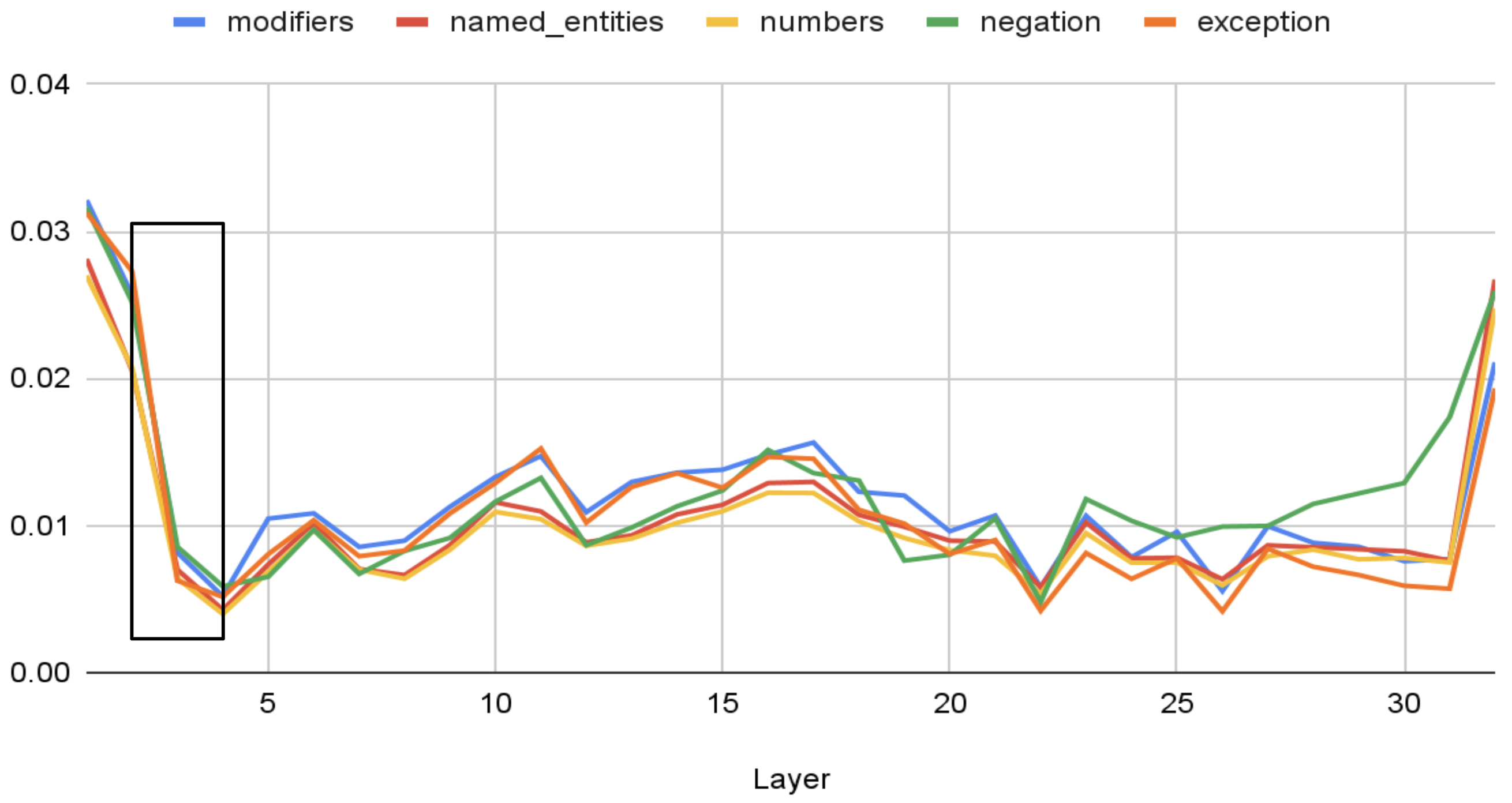}
    \caption{Llama-2-7B-hf (Total Layers-32)}
    \label{fig:Llama7b}
  \end{subfigure}
  \hfill
  \begin{subfigure}[b]{0.48\textwidth}
    \includegraphics[width=\textwidth]{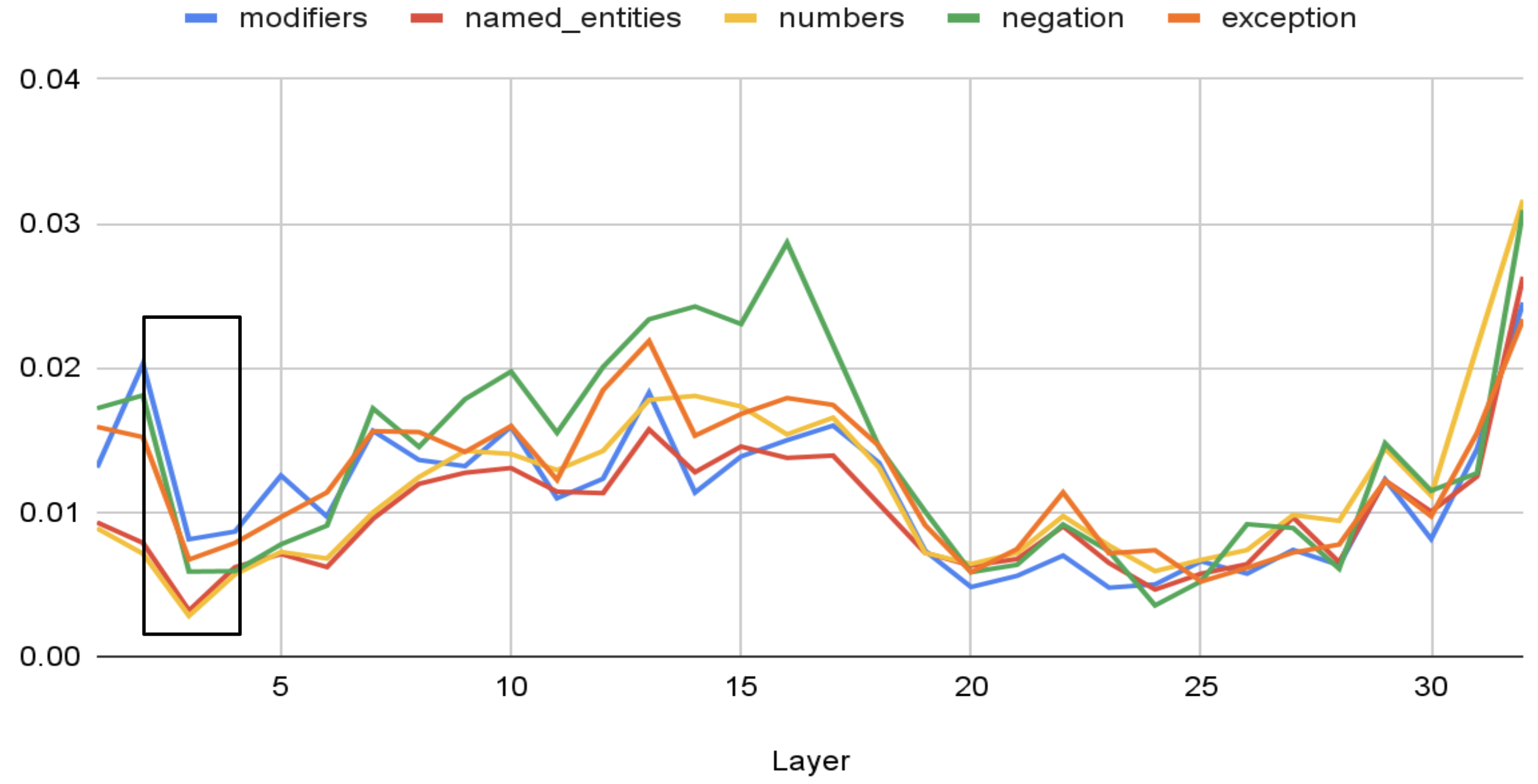}
    \caption{Llama-3.1-8B (Total Layers-32)}
    \label{fig:llama8b}
  \end{subfigure}
  \caption{Standard deviation of symbolic attention across transformer layers for Llama and Gemma models. Early-layer variance (Layers 2–4) indicates symbolic instability that may contribute to hallucination onset. Layers 2–4 are boxed to highlight the earliest emergence and propagation of symbolic misalignment.}
  \label{fig:symbolic_variance}
\end{figure*}

Across Llama-2-7B-hf, Llama-3.1-8B, and Gemma models, a consistent spike in symbolic variance emerges within Layers 2–4 as illustrated in \autoref{fig:symbolic_variance}. This instability is particularly pronounced for negation and exception tokens, which are strongly linked to hallucinated outputs. In Llama-2-7B-hf, symbolic volatility peaks in early layers, stabilizes mid-model, and reappears post Layer 24, especially for negation and modifiers, suggesting delayed symbolic misalignment. Llama-3.1-8B shows a secondary rise between Layers 10–17, again dominated by negation. All models display a late-stage surge at Layer 32 across categories, likely reflecting output dynamics rather than causal instability.

Gemma-2-2B exhibits the strongest early variance, with Layer 2 peaking near 0.17 standard deviation, while deeper layers stabilize around 0.05–0.08. The Gemma-2-9B follows a similar trend with peak variance around 0.15, and both Llama models show nearly identical behavior (~0.16–0.17). Statistical analysis reveals high correlation across models (r > 0.8, p < 0.001), suggesting this is not coincidental but a structural property of transformer architectures \cite{vaswani2017attention, merolla2024layer}.

This concentration of variance in Layers 2–4 is significant because these layers perform critical contextualization tasks: moving beyond token embedding (Layer 1) toward syntax and semantic grounding \cite{rogers2020primer, tenney2019bert}. Instability at this stage propagates through residual connections, compromising later reasoning and answer extraction. Recent findings attribute this to difficulties in modeling short-range dependencies, where unstable attention scores sharpen excessively, producing “attention entropy collapse” \cite{zhai2024stabilizing, hajra2025shortrange}.

Mechanistic studies confirm that factual answering relies on early-layer knowledge enrichment followed by later-layer extraction \cite{jin2024mechanistic}. When early symbolic variance destabilizes retrieval, later layers confidently propagate flawed representations, yielding hallucinations. Notably, scaling (e.g., Gemma-2-9B vs 2B) dampens but does not eliminate the early variance, reinforcing that this instability is intrinsic to transformer design.

These observations suggest that hallucinations are not random generation artifacts but arise from systematic symbolic misalignment in early encoding layers. Symbolic attention variance in Layers 2–4 thus provides a reliable, model-agnostic diagnostic for hallucination localization, enabling proactive detection and targeted architectural interventions.

\begin{table*}[!h]
\centering
\small
\caption{Layer localization of symbolic instability with comparison between attention on symbolic tokens and the highest-attended token at that layer. For each model and property, we report the first instability layer, the symbolic token with its attention value, and the maximum-attended token with its value. For multi-token named entities, the token with the highest attention is shown. "Sym" denotes the symbolic token.}
\scriptsize
\begin{tabular}{p{1.2cm}|p{4.5cm}|p{2.3cm}|p{2.3cm}|p{2.3cm}|p{2.3cm}}
\toprule
\textbf{Symbolic Property} & \textbf{Example} & \textbf{Gemma-2-2B} & \textbf{Gemma-2-9B} & \textbf{Llama-2-7B-hf} & \textbf{Llama-3.1-8B} \\
\midrule
\textbf{Modifiers} & Which is the \textbf{\textit{least}} controversial reform introduced in the past decade? & 
\makecell[l]{\textbf{L3}\\ Sym=least(0.0468)\\ Max=which(0.0925)} & 
\makecell[l]{\textbf{L3}\\ Sym=least(0.0531)\\ Max=which(0.0834)} & 
\makecell[l]{\textbf{L3}\\ Sym=least(0.0134)\\ Max=which(0.0200)} & 
\makecell[l]{\textbf{L3}\\ Sym=least(0.0100)\\ Max=which(0.0160)} \\

\makecell[l]{\textbf{Named}\\ \textbf{Entities}} & What position does Dr. \textbf{\textit{Elena Foster}} hold at \textbf{\textit{NovaGen}} Institute? & 
\makecell[l]{\textbf{L4}\\ Sym=Elena(0.0308)\\ Max=position(0.0938)} & 
\makecell[l]{\textbf{L3}\\ Sym=Elena(0.0530)\\ Max=position(0.0700)} & 
\makecell[l]{\textbf{L3}\\ Sym=Nova(0.0140)\\ Max=.(0.3041)} & 
\makecell[l]{\textbf{L3}\\ Sym=Elena(0.0024)\\ Max=position(0.0080)} \\

\textbf{Numbers} & In what \textbf{\textit{year}} did the Andean Treaty come into force? & 
\makecell[l]{\textbf{L3}\\ Sym=year(0.0385)\\ Max=in(0.0669)} & 
\makecell[l]{\textbf{L3}\\ Sym=year(0.0222)\\ Max=come(0.0751)} & 
\makecell[l]{\textbf{L3}\\ Sym=year(0.0083)\\ Max=in(0.0105)} & 
\makecell[l]{\textbf{L3}\\ Sym=year(0.0116)\\ Max=did(0.0117)} \\

\textbf{Negation} & Which of these countries is \textbf{\textit{not}} a member of the Arctic Council? & 
\makecell[l]{\textbf{L3}\\ Sym=not(0.0451)\\ Max=which(0.1152)} & 
\makecell[l]{\textbf{L3}\\ Sym=not(0.0306)\\ Max=which(0.1054)} & 
\makecell[l]{\textbf{L3}\\ Sym=not(0.0060)\\ Max=which(0.0228)} & 
\makecell[l]{\textbf{L3}\\ Sym=not(0.0060)\\ Max=of(0.0166)} \\

\textbf{Exceptions} & Which mammal lays eggs \textbf{\textit{instead}} of giving birth to live offspring? & 
\makecell[l]{\textbf{L3}\\ Sym=instead(0.0492)\\ Max=giving(0.0681)} & 
\makecell[l]{\textbf{L3}\\ Sym=instead(0.0472)\\ Max=giving(0.0617)} & 
\makecell[l]{\textbf{L3}\\ Sym=instead(0.0052)\\ Max=which(0.0152)} & 
\makecell[l]{\textbf{L3}\\ Sym=instead(0.0060)\\ Max=which(0.0136)} \\
\bottomrule
\end{tabular}
\label{tab:layer_localization}
\end{table*}

Taken together, the results across model scaling, input length, task format, and layer-wise analysis reveal a consistent picture of symbolic hallucination. Variations in architecture, parameter count, or prompt design provide only marginal relief, underscoring that hallucination is not a surface artifact but a structural property of symbolic representation in LLMs. Modifiers, named entities, and negation consistently act as dominant triggers, while early transformer layers (2–4) emerge as the locus of representational instability. This instability propagates forward, yielding confident but factually incorrect outputs.

By reframing symbolic hallucination as a structural misalignment rather than a decoding artifact, our analysis highlights symbolic attention variance as a reliable diagnostic signal—offering a principled basis for proactive detection and architectural interventions.

\section{Conclusion and Future Directions}

This work introduces \textbf{\textsc{SymLoc}}, a symbolic localization framework designed to trace the internal emergence of hallucinations in large language models. By employing symbolic attention variance as a layer-wise metric, we demonstrate that hallucinations often originate from early representational instabilities, particularly when models process symbolic properties such as negation, modifiers, and named entities. Across multiple open-weight models and benchmark datasets, our analysis reveals consistent volatility in early-layer attention, reframing hallucination as a structure-sensitive and symbolically grounded failure. Crucially, \textbf{\textsc{SymLoc}} uncovers both (i) regions where hallucinations are later self-corrected—consistent with prior evidence—and (ii) their initial emergence in early layers, thereby offering an interpretable pathway for detection and correction. The framework’s interpretability arises from leveraging symbolic knowledge to localize and characterize hallucinations within the internal representations of LLMs. Overall, these findings suggest that hallucinations stem less from surface-level decoding or model scale, and more from the internal misrepresentation of symbolic cues.

Building on this framework, future work will explore several directions. First, investigating the specific layers and attention heads associated with symbolic confusion. Second, examining the generalization of symbolic vulnerabilities across diverse model families, including GPT, Mistral, and multilingual or multimodal LLMs. Third, developing symbolic interventions at the prompting level to mitigate ambiguity and hallucination. Finally, extending symbolic localization from the layer level to the token level for finer-grained tracing of hallucination.

In addition, to support further research and benchmarking, we will release the code upon acceptance, along with extended versions of HaluEval and TruthfulQA containing multiple-choice and odd-one-out formats for the community.

\section{Limitations}

While this study sheds light on the symbolic origins of hallucination in large language models, some limitations remain. Our evaluation is limited to two English benchmarks (HaluEval and TruthfulQA), and broader multilingual and domain-specific tests are needed to assess generalizability. Some samples contain overlapping symbolic properties, making it difficult to isolate triggers and potentially inflating cross-category correlations. Finally, we focus only on open-weight transformer models (Gemma and Llama), leaving open whether similar vulnerabilities appear in proprietary or non-transformer architectures.


\bibliographystyle{ACM-Reference-Format}
\bibliography{symbolic_triggers, Manas-Analysis-Figure-4/reference-manas}

\appendix

\end{document}